\pdfoutput=1

\documentclass{elsarticle}

\usepackage[utf8x]{inputenc}
\usepackage{graphicx}
\usepackage{amsbsy}
\usepackage[binary-units]{siunitx}
\usepackage{rotfloat}
\usepackage{booktabs}
\usepackage{multirow}
\usepackage{paralist}
\usepackage{hyperref}
\usepackage[nomain,acronym,nohypertypes={acronym}]{glossaries}
\usepackage[capitalise]{cleveref}

\makeglossaries

\newacronym{adr}{ADR}{active debris removal}
\newacronym{leo}{LEO}{low Earth orbit}
\newacronym{geo}{GEO}{geostationary Earth orbit}
\newacronym{ido}{IDO}{intact derelict object}
\newacronym{ssa}{SSA}{space situational awareness}
\newacronym{cq}{CQ}{competency question}
\newacronym{discos}{DISCOS}{Database and Information System Characterising Objects in Space}
\newacronym{esa}{ESA}{European Space Agency}
\newacronym{owl}{OWL 2}{Web Ontology Language}
\newacronym{w3c}{W3C}{World Wide Web Consortium}
\newacronym{cn}{CN}{criticality number}
\newacronym{lar}{LAR}{launcher adapter ring}
\newacronym{gnc}{GNC}{guidance, navigation \& control}
\newacronym{tracer}{TRACER}{onTology foR ACtive dEbris Removal}
\newacronym{api}{API}{application programming interface}
\newacronym{ric}{RIC}{Robotics Innovation Center}
\newacronym{dfki}{DFKI}{German Research Center for Artificial Intelligence}
\newacronym{rdf}{RDF}{Resource Description Framework}
\newacronym{odp}{ODP}{ontology design pattern}
\newacronym{swrl}{SWRL}{Semantic Web Rule Language}
\newacronym{meo}{MEO}{medium Earth orbit}
\newacronym{gto}{GTO}{GEO transfer orbit}
\newacronym{heo}{HEO}{highly eccentric Earth orbit}
\newacronym{lva}{LVA}{launch vehicle adapter}
\newacronym{asap}{ASAP}{Ariane structure for auxiliary payloads}

\newcommand{\ie}{i.\,e.\ }
\newcommand{\eg}{e.\,g.\ }

\DeclareSIUnit{\degree}{deg}
\renewcommand{\vec}[1]{\boldsymbol{#1}}

\newenvironment{biography}[2]{%
	\normalsize\unitlength 1mm\bigskip\parskip=0pt\par %WB--adjust spacing
	\rule{0pt}{39mm}\vspace{-39mm}\par%   garantees correct page breaking
	\noindent\setbox0\hbox{\includegraphics[width=25mm,height=32mm,clip,keepaspectratio]{#2}} %PW: modern graphics
	\ht0=38mm\count10=\ht0\divide\count10 by\baselineskip%  calculates lines
	\global\hangindent29mm\global\hangafter-\count10%
	\hskip-28.5mm\setbox0\hbox to 28.5mm {\raise-30.5mm\box0\hss}%
	\dp0=0mm\ht0=0mm\box0\noindent\textbf{#1}}{ %AT \bf#1>>\textbf{#1},\rm>>\em %WB
	\par\rm\normalsize}

\hypersetup{
  unicode=false,          % non-Latin characters in Acrobat’s bookmarks
  pdftoolbar=true,        % show Acrobat’s toolbar?
  pdfmenubar=true,        % show Acrobat’s menu?
  pdffitwindow=true,     % window fit to page when opened
  pdftitle={Space Debris Ontology for ADR Capture Methods Selection},    % title
  pdfauthor={Marko Jankovic, Mehmed Yüksel, Mohammad Mohammadzadeh Babr, Francesca Letizia, Vitali Braun},     % author
  pdfsubject={Space Debris Ontology for ADR},   % subject of the document
  pdfcreator={Marko Jankovic},   % creator of the document
  pdfkeywords={space debris, ontology, active debris removal, breakup hazard,
  protégé, owlready2, python}, % list of keywords
  pdfnewwindow=true,      % links in new PDF window
  colorlinks=true,       % false: boxed links; true: colored links
}

\makeatletter
\def\ps@pprintTitle{%
    \let\@oddhead\@empty
    \let\@evenhead\@empty
    \def\@oddfoot{\centerline{\thepage}}%
    \let\@evenfoot\@oddfoot}
\makeatother

\begin{document}

\begin{frontmatter}

% Define name of the journal
\journal{Acta Astronautica}

% Title
\title{Space Debris Ontology for ADR Capture Methods Selection\tnoteref{t1,t2}}

% Title footnotes
\tnotetext[t1]{This document is an accepted version of the manuscript available onlilne at: \url{https://doi.org/10.1016/j.actaastro.2020.03.047}.}
\tnotetext[t2]{\copyright~2020. This manuscript version is made available under the \href{http://creativecommons.org/licenses/by-nc-nd/4.0/}{CC-BY-NC-ND 4.0} license.}

%% Define authors grouped per affiliation:
% DFKI-RIC
\author{Marko Jankovic\corref{correspondingauthor}}
\cortext[correspondingauthor]{Corresponding author.}
\ead{marko.jankovic@dfki.de}
\author{Mehmed Yüksel}
\author{Mohammad Mohammadzadeh Babr}
\address{Robotics Innovation Center (RIC)--DFKI GmbH and University of Bremen,\\
Robert-Hooke-Straße 1, 28359 Bremen, Germany}
% ESA-ESOC
\author{Francesca Letizia}
\author{Vitali Braun}
\address{IMS Space Consultancy for the European Space Operation Center (ESOC)--ESA,\\Robert-Bosch-Straße 5,
64293 Darmstadt, Germany}

%%%%%%%%%%%%%%%%%%%%%%%%%%%%% ABSTRACT & KEYWORDS %%%%%%%%%%%%%%%%%%%%%%%%%%%%%%

% Define the abstract
\begin{abstract} \label{sec:abstract}

Studies have concluded that \gls{adr} of the existing in-orbit mass is necessary.
However, the quest for an optimal solution does not have a unique answer and the available data often lacks coherence.
To improve this situation, modern knowledge representation techniques, that have been shaping the World Wide Web, medicine and pharmacy, should be employed.
Prior efforts in the domain of space debris have only focused onto space situational awareness, neglecting \gls{adr}\@.
To bridge this gap we present a domain-ontology of intact derelict objects, \ie payloads and rocket bodies, for \gls{adr} capture methods selection.
The ontology is defined on a minimal set of physical, dynamical and statistical parameters of a target object.
The practicality and validity of the ontology are demonstrated by applying it onto a database of 30 representative objects, built by combining structured and unstructured data from publicly available sources.
The analysis of results proves the ontology capable of inferring the most suited \gls{adr} capture methods for considered objects.
Furthermore, it confirms its ability to handle the input data from different sources transparently, minimizing user input.
The developed ontology provides an initial step towards a more comprehensive knowledge representation framework meant to improve data management and knowledge discovery in the domain of space debris. Furthermore, it provides a tool that should make the initial planning of future \gls{adr} missions simpler yet more systematic.

\end{abstract}

%Define the keywords

\end{frontmatter}

%\linenumbers

%%%%%%%%%%%%%%%%%%%%%%%%%%%%%%% ARTICLE CONTENT %%%%%%%%%%%%%%%%%%%%%%%%%%%%%%%%

% Reset all acronyms
\glsresetall

% Define individual sections of the article
%!TEX root = main.tex

\section{Introduction} \label{sec:introduction}

The space infrastructure has become an integral part of the human society.
Even the most mundane services, such as navigation, weather forecast,
television and banking, are nowadays supported by the on-orbit
infrastructure.
It is therefore understandable that after more than 60 years of space activities, which resulted in more than \num{5480} launches and \num{520} fragmentations~\citep{discos2019, ESASDO2019}, there is a growing concern of the society at large over the space debris issue, that places vital services in a constant danger of collision.
This situation is being even more exacerbated with the rise of  mega-constellations, from commercial companies such as OneWeb, SpaceX and Amazon, that plan to deploy more than \num{53000} new satellites in \gls{leo}.
However, dealing with the issue of space debris is at present a challenging task, considering the amount of uncertainty surrounding it~\citep{IAA2017}.
In fact, currently, we can only reliably detect and catalog objects that
are larger than \SIrange{5}{10}{\centi\meter} in \gls{leo} and \SIrange{30}{100}{\centi\meter} in \gls{geo}~\citep{ESA2018}.
The population of smaller objects can be estimated only by dedicated models and validated through retrieved surfaces, which were exposed to the space debris environment.
In addition, although certain orbital regions, such as \gls{leo}, have theoretically exceeded the critical density of objects that would ensure the onset of the Kessler syndrome\footnote{A self-sustaining collision process that would increase the number of on-orbit objects exponentially, due to series of cascading collisions among objects of the existing population~\citep{Kessler1978}.}, it is uncertain when and in which measure exactly will this phenomenon appear on-orbit~\citep{IAA2017}.
Furthermore, the currently implemented space debris mitigation guidelines~\citep{IADC2007} have proven to be necessary but insufficient activities
to manage the risk posed by the space debris and therefore maintain a stable space debris environment~\citep{IAA2017}.
The only way to permanently stabilize the current environment and prevent the onset of the Kessler syndrome consists of actively removing the existing large sources of debris, \ie \glspl{ido}\footnote{Defined as objects having a size \SI{> 1}{\meter} and \SI{> 2}{\meter} in \gls{leo} and \gls{geo}, respectively~\citep{IAA2017}.}, via remediation activities. This way future collisions between large objects can be prevented and with them the generation of fragmentation debris that are more difficult to track and, currently, neither practical nor economically feasible to remove actively~\citep{IAA2017,Kaplan2010a}.

Among the currently investigated remediation activities, \gls{adr} is perceived as the only one able to permanently reduce the number of \glspl{ido}, at the expense of difficult mission planning, execution and necessity to act over a period of decades.
Among all the phases of a generic \gls{adr} mission, the capture phase emerges as one of the most challenging ones (along with the close-range rendezvous), since, to best of our knowledge, no spacecraft has ever captured a completely uncooperative target.
It consists of actions, performed by a chaser spacecraft, to capture a target, stabilize the compound and prepare it for
disposal (\ie de- or re-orbit). Based on the type of ``contact'' needed to successfully capture a target, the capture approaches currently being researched can be grouped into: contact-based or contactless methods.
In fact, while the former include technologies requiring a physical
contact with a target, the latter are able to achieve a ``capture'' by actively
controlling the attitude of a target from a stand-off distance. Contact-based
methods can be further divided into~\citep{Jankovic2016b}:
\begin{description}
	\item [{robotics-based\ systems}] employing robotic devices (\eg a
	manipulator or clamp/tentacle) to capture a target and stabilize the
	compound
	\item [{tether-based\ systems}] operating a tethered net or harpoon to
	capture a target from a stand-off distance.
\end{description}

Contactless methods can be likewise further grouped into~\citep{Jankovic2016b}:
\begin{description}
	\item [{plume\ impingement-based\ systems}] using an ion- or inert gas-based
	engine to create a plume of particles in front of a target to reduce its
	momentum
	and therefore achieve a ``capture'' within a predefined volume of space, from
	a stand-off distance
	\item [{ablation\ systems}] applying a concentrated source of electromagnetic
	radiation (\eg visible light) to ablate
	the surface of a target, thus generating a small, but constant thrust
	opposite to the direction of the applied radiation
	\item [{electromagnetic-based\ systems}] (\eg eddy brakes, electrostatic
	tractors) exploiting electromagnetic or electrostatic forces to envelop
	a target in a magnetic or electric vector field, respectively, and
	generate necessary dissipative forces to ``capture'' it.
\end{description}

Every method has his own advantages and disadvantages but there is unfortunately not one that can tackle all possible targets.
Moreover, even considering one specific target, it is
not guaranteed to be able to easily identify its most suited method(s), as
demonstrated by the e.Deorbit study~\citep{Biesbroek2013}.
Furthermore, the available data about cataloged objects often lacks coherence
and structure, thus hindering data sharing, collaboration and ultimately
decision-making.
In fact, it might be argued that the current space debris issue is plagued
by the information paradox~\citep{Orman2017} where we appear to be: ``... drowning in information but starved for knowledge''~\citep{Naisbitt1982}.

In this context, we present a knowledge representation framework for the characterization
of space debris for \gls{adr} capture methods selection in the form of a domain-ontology.
The latter is defined in our research as a method to model a field of discourse by explicitly defining the domain concepts, relationships among them, their properties and
restrictions~\citep{NatalyaF.Noy2001}.
This way, standardized, machine-interpretable vocabulary of characteristics of
cataloged objects, and relations among them, can be used by ADR researchers to
analyze the domain and infer new knowledge, making the initial \gls{adr}
mission planning easier yet more systematic.
The nature of our results is both theoretical and applied, since not only the
methodology used to develop the domain-ontology is described, but also
workflows used for its software implementation and usage.
With respect to the existing state-of-the-art~\citep{Wilkins2013,Frueh2013,Cox2016, Rovetto2016, Furfaro2016, Liu2016, LeMay2018}, focused mainly on \gls{ssa}, the ontology covers specifically the domain of \gls{adr} (or more specifically the domain of \gls{adr} capture methods) and establishes the minimal number of parameters needed to identify the most suited capture method(s) for a specific target.
Additionally, it describes a method to handle the input of data from an existing database of cataloged objects, an aspect often overlooked by the current state-of-the-art.

The remainder of the article is structured as follows:
\cref{sec:methodology} introduces the overall methodology and core terminology of the ontology.
In \cref{sec:ontology_implementation} its software implementation and development workflows are described.
\cref{sec:ontology_application} presents results of its application onto a database of representative objects.
The analysis of those results, along with strengths and weaknesses of the ontology, are discussed in \cref{sec:analysis_and_discussion}.
Finally, \cref{sec:conclusions} is dedicated to the concluding remarks and
recommendations for future works.
%!TEX root = main.tex

\section{Methodology and Core Terminology} \label{sec:methodology}

The overall methodology of this work consists of the development of an ontology
for \gls{adr} capture methods inference, based on physical and dynamical
properties of potential targets. In order to do so, firstly the precise domain
of interest and scope of the ontology are defined. Then, main classes and their
hierarchy are determined. Finally, axioms allowing classification of \gls{adr}
capture methods, associated with potential targets, are specified, based on the
statistical analysis of on-orbit fragmentations and state-of-the-art capture
methods characteristics.

\subsection{Domain of interest and scope}\label{subsec:ontology-domain-scope}

The methodology used to develop the ontology is primarily
a ``Simple Knowledge-Engineering Methodology'' \citep{NatalyaF.Noy2001}, with some hints taken from the Unified Process for ONtology (UPON) method \citep{DeNicola2009}.
It consists of an iterative approach where at first a rough draft
of the ontology is developed, starting from the initially defined requirements set and
lexicon.
Then, the lexicon is transformed into a glossary.
Subsequently, the ontology is revised and enriched based on the analysis of
the existing draft. After several iterations of the previous step, the ontology
is finally formalized and tested using a representative set of data against the
initial set of requirements.

The domain of the developed ontology is that of \glspl{ido} (\ie payloads and rocket bodies) and \gls{adr} capture methods. Therefore, at least
in its current version, it does not cover the domain of de-orbit
technologies, although it is formulated with that domain in
mind. The scope of the ontology is to create a standardized framework for collection,
storage and sharing of characteristics of \glspl{ido} for \gls{adr}, by leveraging
attributes of modern knowledge representation techniques. This
way not only a method for efficient storage of complex information is provided
but also a platform for further analysis of the domain knowledge from
the existing data. One such analysis is provided with the ontology
and consists of inferring, for a specific object, the most suited
ADR capture method(s) and provide a human-readable explanation of the inference.

\begin{figure}
	\begin{centering}
		\includegraphics[width=.9\linewidth]{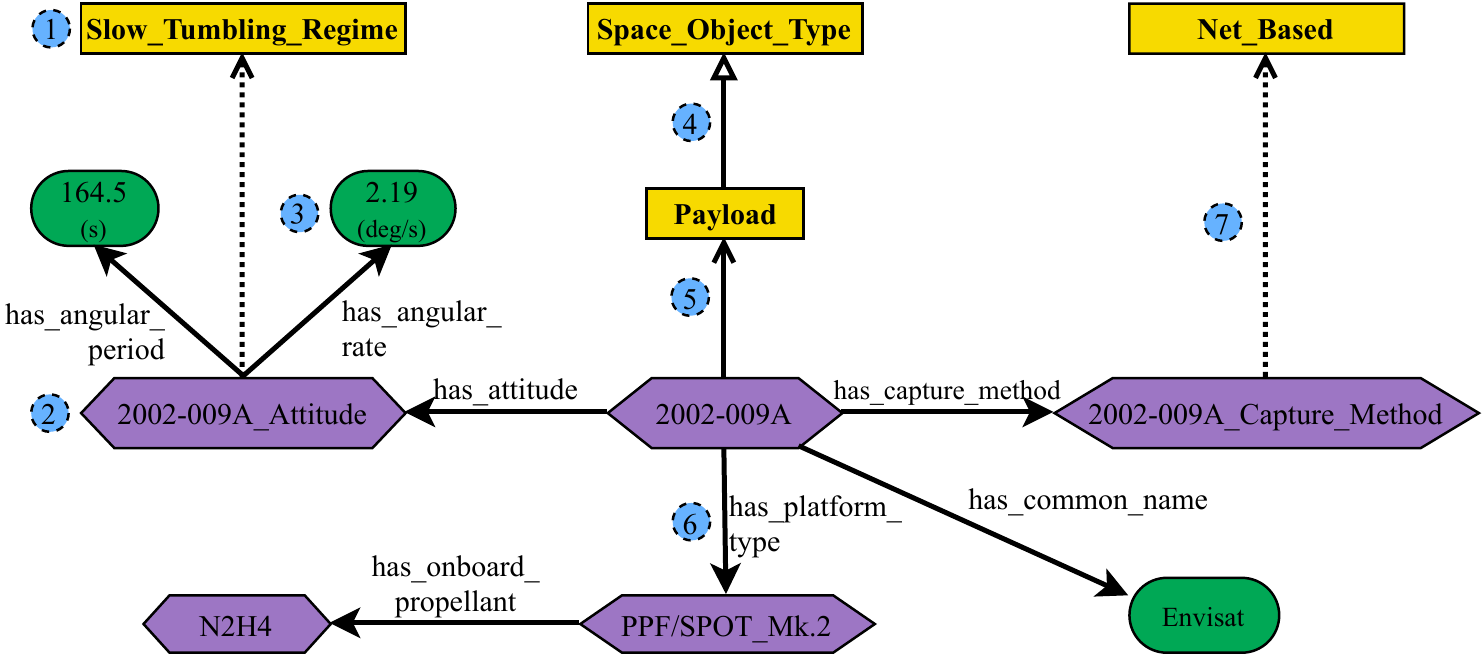}
		\par\end{centering}
	\caption{\label{fig:ontology-example}
		Exemplary diagram of the developed ontology. The conventions used in
		the diagram are the following:
		\protect\begin{inparaenum}[(1)]
			\protect\item yellow squares represent classes,
			\protect\item purple hexagons represent individuals,
			\protect\item green rounded squares depict numerical or string values,
			\protect\item closed hollow arrows portray subclass or subproperty relations,
			\protect\item opened arrows illustrate ``type of'' relations,
			\protect\item semi closed solid arrows depict properties of individuals,
			\protect\item dashed arrows portray inferred axioms.
		\protect\end{inparaenum}}
\end{figure}

The \glspl{cq} the ontology should provide answers
to are identified as the following:
\begin{description}
	\item [\gls{cq}1] How could a domain knowledge about \glspl{ido} be captured in a
	standardized, formal, machine-interpretable way useful to \gls{adr}\@?
	\item [\gls{cq}2] What are the minimum parameters needed to characterize an
	\gls{ido} for an \gls{adr} capture phase?
	\item [\gls{cq}3] How can the degree of hazard of an \gls{ido} to an \gls{adr} capture
	phase be represented?
	\item [\gls{cq}4] How could the most suited \gls{adr} capture method be inferred?
	\item [\gls{cq}5] How could the input of data into the ontology be simplified
	and made compatible with an existing space debris catalog, such as the \gls{discos} of the \gls{esa}~\citep{discos2019}?
\end{description}

The intended users of the ontology are space debris domain experts,
ADR mission planners and decision makers, that should have at their
disposal a standardized way for data collection, storage and access
of complex domain knowledge, such as that of \glspl{ido} and \gls{adr}\@.
By using such a framework the parameter space of each object
can be kept hidden from the user, as much as possible, and usage of a semantic
reasoner allows for queries and knowledge inference from the existing data.

\subsection{Main classes and class hierarchy\label{subsec:main-classes}}

Among the existing knowledge representation languages, this work
considers the \gls{owl}, developed by the \gls{w3c} for the Semantic Web.
Ontologies developed with \gls{owl} store the information about the domain of interest into Semantic Web documents, capable of representing classes, properties, individuals
and relationship among them (see \cref{fig:ontology-example}) \citep{owl2-2012}.

Classes are the main building blocks of an \gls{owl} ontology and their
identification generally involves analyzing and extracting terms from
the existing documentation, technical manuals, standards and/or similar
ontologies \citep{NatalyaF.Noy2001,DeNicola2009}. This was done
in our work by extracting the most important terms from the preceding
taxonomy of \gls{leo} space debris \citep{Jankovic2016b,Jankovic2018c},
whose purpose was similar, albeit more limited, due to the inherent
constraints of the used knowledge representation method (\ie a taxonomy).

Classes identified as sufficient to characterize an \gls{ido}, in an unambiguous
manner for a capture maneuver, and included within the developed ontology
are: \textsf{Attitude\_Regime, Onboard\_Propellant, Space\_Object\_Type,
	Breakup\_Criticality} and \textsf{ ADR\_Capture\_Method}. Additional
classes, \ie \textsf{Launch\_Vehicle}, \textsf{Orbital\_Regime,}
\textsf{Spacecraft\_Platform} and \textsf{Stage\_Type,} are
introduced for completeness, however are deemed nonessential for the purpose
of this work. Inclusion of further classes although possible, was excluded in
order to keep the ontology simple and clutter free.

\begin{figure}
	\centering{}\includegraphics[height=0.9\textheight]{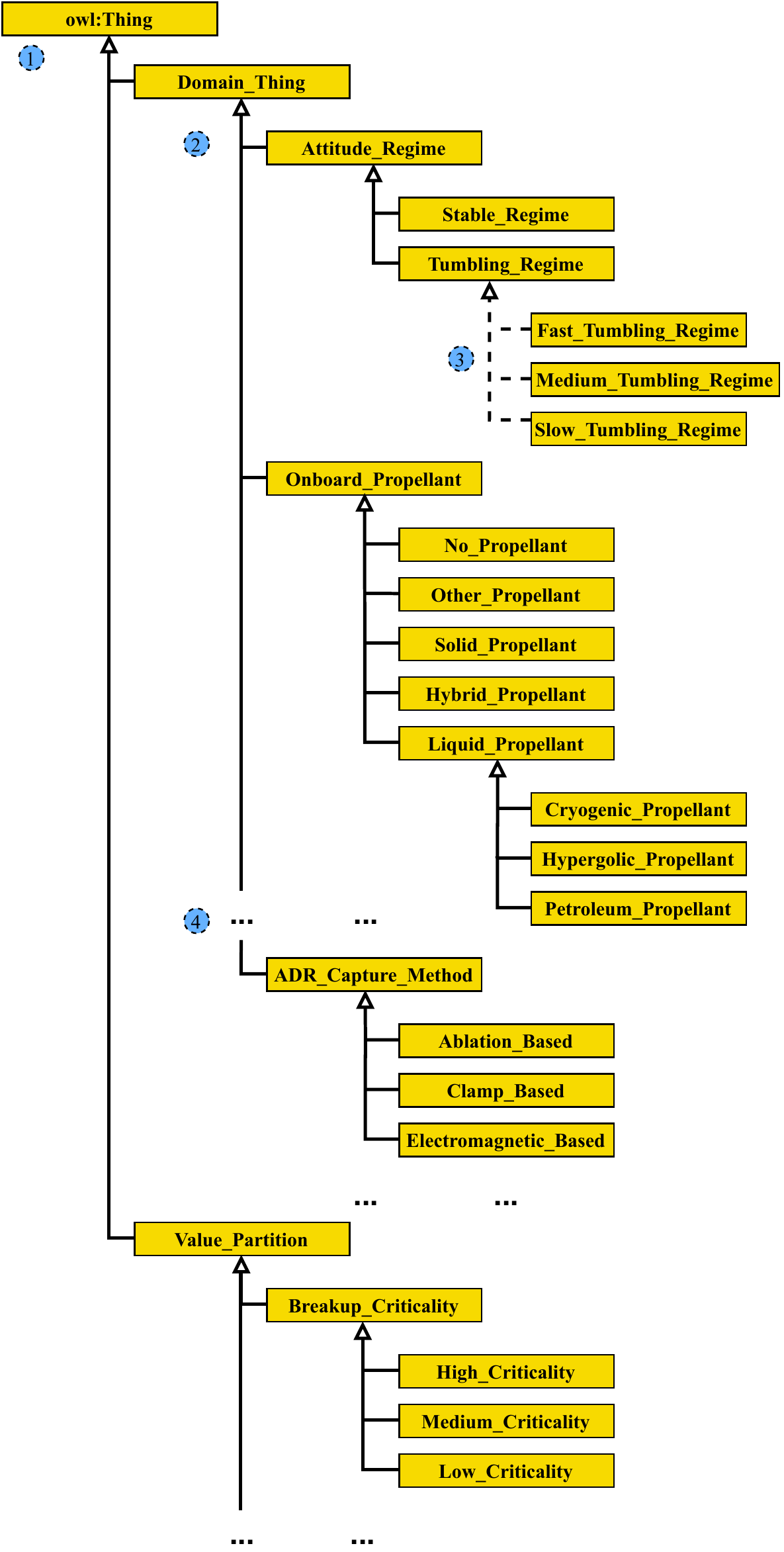}
	\caption{\label{fig:hiererchy-example}Extract of the class hierarchy from the
	developed ontology. The conventions used in the diagram are the following:
	\protect\begin{inparaenum}[(1)]
		\protect\item yellow squares represent classes,
		\protect\item closed hollow arrows portray subclass relations,
		\protect\item dashed arrows portray inferred axioms,
		\protect\item three dots indicate an omission of the existing structure.
	\protect\end{inparaenum}}
\end{figure}

The hierarchy of those classes consists, as illustrated in \cref{fig:hiererchy-example},
of three levels, each specifying a domain concept
into more detail. Starting from the top of \cref{fig:hiererchy-example}, the \textsf{Attitude\_Regime}
class characterizes instances describing the attitude state of an
object. It contains two additional layers specifying the exact type
of attitude state of an object, which in our study is confined
to either being stable or tumbling (\ie \textsf{Stable\_} or
\textsf{Tumbling\_Regime}). The former is defined as a state where the angular
velocity of a target object is equal to zero \si{\degree\per\second} \ie $\vec{\omega}_{t}=\SI{0}{\degree\per\second}$.
The latter is instead further detailed into fast, medium and slow tumbling
regimes (\ie\textsf{Fast\_}, \textsf{Medium\_} and \textsf{Slow\_Tumbling\_Regime} subclasses), each defined
as the angular rate of an object being: $0<\vec{\omega}_{t}<\SI{5}{\degree\per\second}$,
$5\leq\vec{\omega}_{t}<\SI{18}{\degree\per\second}$, $18\leq\vec{\omega}_{t}<\infty\,\si{\degree\per\second}$,
respectively~\citep{Jankovic2016b}.

The \textsf{Onboard\_Propellant} class defines the propellant type
of the main propulsion system of an \gls{ido}, or better of its platform
(\ie bus or propulsion unit, in case of a payload or rocket body,
respectively).
Based on the state of matter of the oxidizer and fuel used by the propulsion
system, a distinction is made between \textsf{Liquid\_}, \textsf{Solid\_}
and \textsf{Hybrid\_Propellant} \citep{Braeunig2008}. Two
additional classes are added for completeness, \ie
\textsf{No\_Propellant} and \textsf{Other\_Propellant}, in case of spacecraft
having no propulsion system or having an onboard propellant that cannot be
classified within one of the previous classes (\eg butane, xenon, etc.).
Lastly, the \textsf{Liquid\_Propellant} class, is further subdivided
into \textsf{Cryogenic\_}, \textsf{Hypergolic\_} and \textsf{Petroleum\_Propellant}
subclasses, based on the specific oxidizer-fuel mixture used by the propulsion
system, as defined in \citep{Braeunig2008}.

The type of \gls{ido} is specified with the \textsf{Space\_Object\_Type}
class, which is further subdivided
in the two most relevant object types identified
in this work, \ie \textsf{Payload} and \textsf{Rocket\_Body}, defined as
in \citep{ESASDO2019}.

The remaining main classes, \ie \textsf{Breakup\_Criticality} and
\textsf{ADR\_Capture\_Method}, represent the core of the developed ontology, as
they are instrumental in the ADR capture method inference.

The \textsf{Breakup\_Criticality} class establishes the breakup hazard
of an object, due to its inherent probability of breakup, as a consequence of
variety of causes, as defined in \citep{ESASDO2019}.

The \textsf{ADR\_Capture\_Method} class instead defines
the suitability of an ADR capture technology, described in \cref{sec:introduction},
to capture an object based on its breakup criticality and level of
uncooperativeness. The latter is defined as the degree of difficulty
to capture a particular object due to its dynamical and physical properties
\citep{Jankovic2016b}.

More details about both of these concepts and how they influence the
classification of capture methods are outlined in the next two subsections.

\subsection{Breakup criticality definition}\label{subsec:breakup-criticality-definition}

In this work, the breakup criticality of an object is formalized in alignment
with \gls{esa}['s] standard on failure modes, effects (and criticality) analysis
(FMEA/FMECA) \citep{ECSS2009a}. Therefore, it is defined as a combination of a
severity and probability of occurrence of a fragmentation event (\ie the
associated severity and probability numbers, identified hereafter with symbols
SN and PN, respectively), having the worst possible consequences for a
capture maneuver (see \cref{tab:criticality-matrix}).
Consequently, a distinction is made not only between different object types,
\eg rocket bodies and payloads, but also between non-passivated
and passivated objects \citep{Jankovic2016b}.

\begin{table}
	\caption{\label{tab:criticality-matrix}Criticality matrix (adapted from Table~5-3 of \citep{ECSS2009a}).}

	\centering{}%
	\begin{tabular*}{0.9\textwidth}{@{\extracolsep{\fill}}>{\raggedright}p{0.15\columnwidth}>{\raggedright}p{0.1\columnwidth}cccc}
		\toprule
		\multirow{4}{0.2\columnwidth}{Severity\\
			Level} & \multirow{4}{0.1\columnwidth}{Severity \\
			Number \\
			(SN)} & \multicolumn{4}{c}{Probability Limits}\tabularnewline
		\cmidrule{3-6} \cmidrule{4-6} \cmidrule{5-6} \cmidrule{6-6}
		&  & \num{\leq e-4} & \num{\leq e-2} & \num{\leq e-1} & \num{> e-1}\tabularnewline
		\cmidrule{3-6} \cmidrule{4-6} \cmidrule{5-6} \cmidrule{6-6}
		&  & \multicolumn{4}{c}{Probability Number (PN)}\tabularnewline
		\cmidrule{3-6} \cmidrule{4-6} \cmidrule{5-6} \cmidrule{6-6}
		&  & \textbf{1} & \textbf{2} & \textbf{3} & \textbf{4}\tabularnewline
		\midrule
		\textbf{Catastrophic} & \textbf{4} & 4 & 8 & 12 & 16\tabularnewline
		\textbf{Critical} & \textbf{3} & 3 & 6 & 9 & 12\tabularnewline
		\textbf{Major} & \textbf{2} & 2 & 4 & 6 & 8\tabularnewline
		\textbf{Negligible} & \textbf{1} & 1 & 2 & 3 & 4\tabularnewline
		\bottomrule
	\end{tabular*}
\end{table}

An object is characterized with a high criticality level
(\ie high \gls{cn}) and is considered as critical for
capture when one of the following conditions is fulfilled
\citep{Jankovic2016b}:

\begin{itemize}
	\item the severity of its worst fragmentation event is defined as
	catastrophic, \ie its $ \text{SN} = 4 $
	\item the criticality number of its worst fragmentation event is greater or equal to eight, \ie its $ \text{CN} \geq 8 $ (see
	\cref{tab:criticality-matrix}).
\end{itemize}

In these cases, any close contact with a target is to be avoided and
only methods capable of achieving a capture from a significant stand-off
distance (\eg \SI{>=50}{\meter}) should be considered.
Moreover, in these cases special care should
be exerted during the capture and stabilization maneuvers to avoid
shocks and sources of sparks that might trigger a catastrophic breakup
\citep{Jankovic2016b}.

The remaining criticality levels are defined in this work as:

\begin{description}
	\item [medium] if the criticality number of its worst fragmentation event
	is equal to six, \ie its $ \text{CN} = 6 $
	\item [low] if the severity level of its worst fragmentation event is
	considered negligible, or its criticality number is lower or
	equal to four, \ie its $ \text{CN} \leq 4 $.
\end{description}

In these cases, several capture methods can be employed, depending not only on
the CN but also on physical and dynamical properties of an object, such as its type, attitude regime, etc., as it will be described in the next subsection.

The worst possible breakup event of rocket bodies was identified as either an anomalous or propulsion event, depending on whether the spacecraft were passivated or not, based on a statistical analysis of related fragmentations extracted from DISCOS.
The severity of those events, distinguished by the onboard propellant type, is listed in \cref{tab:severity-rocket-bodies}.
The ``No. of fragments'' column details the observed median number of fragments
per breakup event.
The threshold of 1.05 years in the “Parent objects” column indicates the orbit age of an object (from launch) after which a propulsion related event can be expected to be statistically less severe, possibly due to a depletion/venting of most of the stored on-board propellant.

The probability of occurrence of those events was estimated using the survival analysis that was successfully employed to estimate the reliability of spacecraft from incomplete data \citep{Saleh2011a}.
For this purpose, the data was collected and processed from \gls{discos},
containing information about all large rocket bodies (not related
to any manned mission) from October 1957 till July 2019, resulting
in a total of \num{5185} cataloged objects and \num{15555} observed
(breakup and censored) events. Censoring occurs either because:
\begin{inparaenum}[a)]
	\item an object has reentered,
	\item the reentry date of an object is beyond the observational window (in our case July 2019),
	\item the reentry date of an object is unknown.
\end{inparaenum}
Furthermore, during the analysis of a
particular breakup, all other breakups form an additional source of
censoring that needs to be accounted for \citep{Saleh2011a}.
The survival function $S(t)$ of the population of objects
within the database, and therefore the probability function
from which it is derived, $P(t)=1-S(t)$, was estimated using the non-parametric Kaplan-Meier estimator \citep{Kaplan1958}.
While other methods exist, this one was chosen for its accuracy, as it does not fit any pre-defined distribution and is based on the actual data \citep{Saleh2011a}.

\begin{table}
	\caption{\label{tab:severity-rocket-bodies}Severity numbers of the worst fragmentation events of rocket bodies in terms of the medium number of fragments (RBs = rocket bodies; N/A = not applicable).}

	\centering{}%
	\begin{tabular*}{1\columnwidth}{@{\extracolsep{\fill}}>{\raggedright}m{0.05\columnwidth}>{\centering}m{0.15\columnwidth}>{\centering}m{0.15\columnwidth}>{\centering}m{0.1\columnwidth}>{\centering}m{0.35\columnwidth}}
		\toprule
		\multirow{1}{0.05\textwidth}{SN} & Breakup class & Propellant type & No.\ of \\
		fragments & Parent objects\tabularnewline
		\midrule
		\midrule
		4 & Propulsion & All & \num{\geq 200} & Non-passivated RBs with

		orbit age \num{\leq 1.05} years\tabularnewline
		\midrule
		3 & Propulsion & Hypergolic & 79.5 & \multirow{3}{0.35\textwidth}{\centering{}Non-passivated RBs with orbit age
			\num{> 1.05} years}\tabularnewline
		\cmidrule{1-4} \cmidrule{2-4} \cmidrule{3-4} \cmidrule{4-4}
		2 & Propulsion & Cryogenic & 7.5 & \tabularnewline
		\cmidrule{1-4} \cmidrule{2-4} \cmidrule{3-4} \cmidrule{4-4}
		1 & Propulsion & Petroleum \& Solid & 2.5 & \tabularnewline
		\midrule
		1 & Anomalous & N/A & 1 & Passivated RBs\tabularnewline
		\bottomrule
	\end{tabular*}
\end{table}

\begin{figure}
	\begin{centering}
		\includegraphics[width=.9\linewidth]{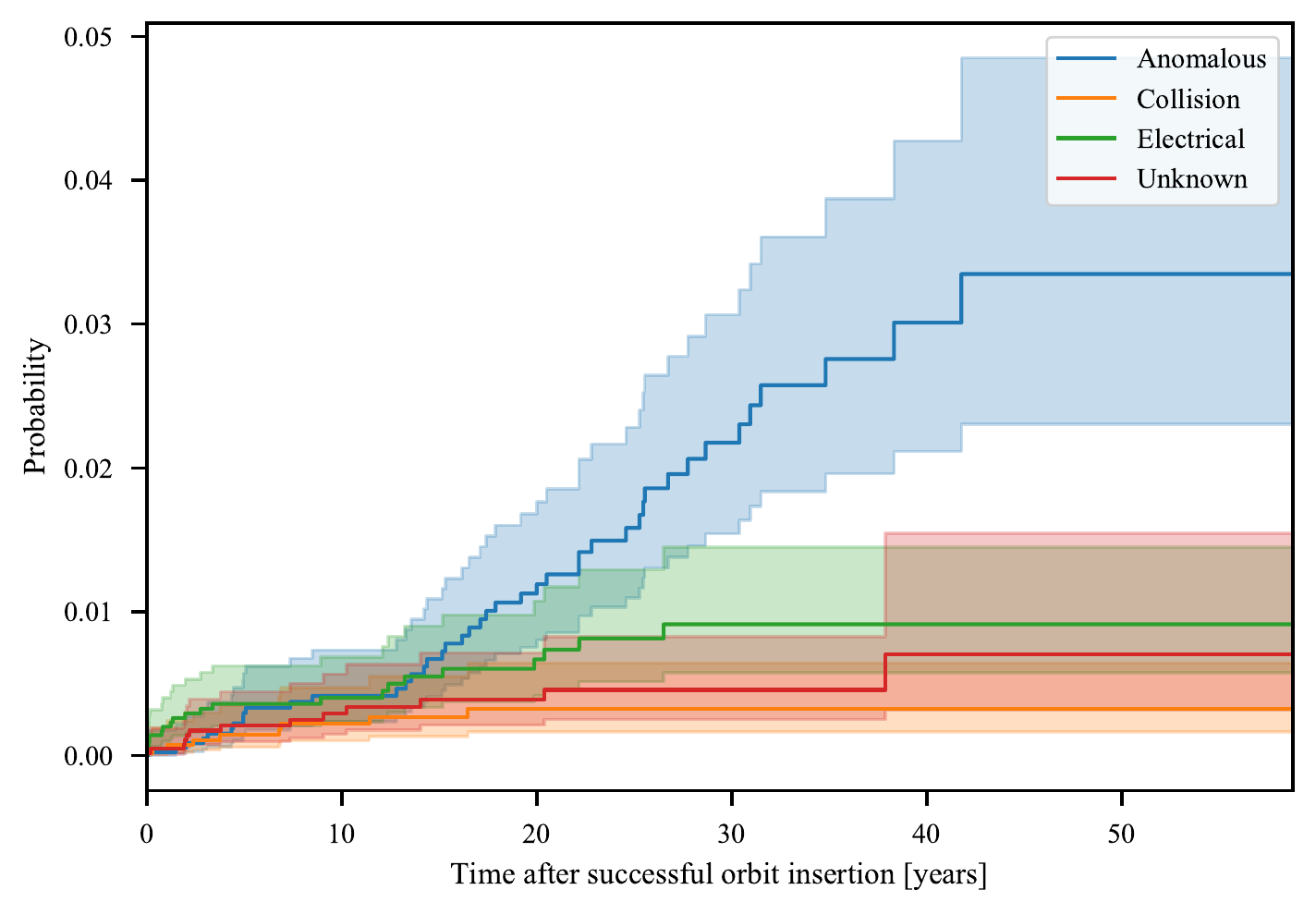}
		\par\end{centering}
	\caption{\label{fig:probability-payloads}Breakup probabilities (with \SI{95}{\percent} confidence intervals) of large, unmanned payloads due to anomalous, collision, electrical and unknown causes calculated using the Kaplan-Meier estimator.}
\end{figure}

%TODO: clarify the 1.05 years thershold
In case of payloads, the worst possible breakup event of non-passivated spacecraft was found to be a combination of anomalous and electrical events. The combination of anomalous, collision and unknown events was instead taken into account in case of passivated spacecraft. The severities of those events were both assigned the major severity level, \ie $ \text{SN} = 2 $, due to their similar median number of generated fragments.
The probability of occurrence of mentioned events was estimated, as in case of rocket bodies, using the Kaplan-Meier estimator applied on data consisting of \num{5002} payloads and \num{20008}
observed (breakup and censored) events.
However, since in both cases the dominant probability function was that of anomalous events (see~\cref{fig:probability-payloads}), no significant difference was found between the derived probabilities.

\subsection{ADR capture method selection}\label{subsec:capture-method-selection}

In this work, the selection of the most suited ADR capture method(s) for a specific target is determined via the evaluation of its previously defined
breakup hazard and degree of uncooperativeness. This
way, not only the safety of the capture maneuver is considered during
the evaluation, but also its degree of difficulty, which is manifested
by the physical and dynamical properties of an object.

The characteristics of an object used to define its degree of uncooperativeness
are:
\begin{inparaenum}[1)]
	\item the angular rate,
	\item existence of a dedicated grapple feature,
	\item type of material of the capture interface,
	\item mechanical clearance of the capture interface.
\end{inparaenum}
The definition of those characteristics is summarized in
\cref{tab:uncoop-characteristics} and detailed hereafter based on the perceived capabilities of the current state-of-the-art capture methods. Therefore, the threshold values used to define them in this article are provided as default ones. Users are encouraged to modify them or even introduce new ones using software tools presented in the next section.

\begin{table*}
	\caption{\label{tab:uncoop-characteristics}Summary of characteristics of a target used to determine its degree of uncooperativeness ($ \omega_{t}= $ angular velocity of a target; $ A= $ area centered on the capture interface of a target) (adapted from Table~6 of \citep{Jankovic2018c}).}

	\centering{}%
	\begin{tabular*}{1\textwidth}{@{\extracolsep{\fill}}>{\raggedright}p{0.3\textwidth}>{\raggedright}p{0.7\textwidth}}
		\toprule
		Characteristic & Definition\tabularnewline
		\midrule
		Angular rate & Stable: $\vec{\omega}_{t}=\SI{0}{\degree\per\second}$

		Slow tumbling: $0<\vec{\omega}_{t}<\SI{5}{\degree\per\second}$

		Medium tumbling: $5\leq\vec{\omega}_{t}<\SI{18}{\degree\per\second}$

		Fast tumbling: $18\leq\vec{\omega}_{t}<\infty\,\si{\degree\per\second}$\tabularnewline
		\midrule
		Grapple feature existence & True: \textit{``dedicated'' grapple feature exists}

		False: \textit{``dedicated'' grapple feature does not exist}\tabularnewline
		\midrule
		Capture interface material & Isotropic: \textit{\eg metal, ceramics, polymer}

		Anisotropic: \textit{\eg composite materials}\tabularnewline
		\midrule
		Capture interface clearance & Narrow: $A<\SI{0.28}{\meter\squared}$

		Broad: $A\geq\SI{0.28}{\meter\squared}$\tabularnewline
		\bottomrule
	\end{tabular*}
\end{table*}

The \textbf{angular rate} is expressed in the ontology as a data property and describes the attitude state of an object.
The thresholds defining the tumbling states, visible in
\cref{tab:uncoop-characteristics}, are chosen based on:
\begin{inparaenum}[a)]
	\item the maximum value of the relative angular rate that a state-of-the-art robotic
	manipulator should be able cope with (\ie
	\SI{5}{\degree\per\second}) \citep{Castronuovo2011,
	Bonnal2013a},
	\item the value of the relative angular rate above which any
	synchronization effort would be considered very difficult (\ie \SI{18}{\degree\per\second})
	\citep{Matsumoto2002}.
\end{inparaenum}
Consequently, objects having angular rates greater or equal to
\SI{18}{\degree\per\second} are assigned a high degree of
uncooperativeness and should be captured only with contactless ``capture'' methods \citep{Jankovic2016b}.

The \textbf{grapple feature existence} is defined as a Boolean data
type property.
A ``dedicated'' grapple features, is identified in this work as a surface
feature, with a regular enough geometry (\eg \gls{lar}, common on many spacecraft), that can be easily grappled.
Otherwise, a capture has necessarily to be performed on some
other feature, not envisioned to be grappled, or even a surface \citep{Jankovic2016b}.
Therefore, the existence of a grapple feature is considered advantageous for the capture maneuver since it can be approximated with a more common berthing operation, commonly used in the context of loading/unloading of cargo from the International Space Station.

The \textbf{capture interface material} is defined as a string data
type property and reflects the versatility (and reliability) of a capture
method.
Considering the directional dependent mechanical properties of an isotropic type
of material, it is associated with capture methods capable of
distributing applied contact forces (such as clamp or net-based
methods). An anisotropic material is, on the other hand, expected to be
able to withstand concentrated loads, independent of their direction,
which is why it is instead correlated with capture methods that
are anticipated to exert such forces
(such as manipulators or harpoon-based methods) \citep{Jankovic2016b}.

The \textbf{capture interface clearance} is expressed as a
string data type. It reflects the overall complexity of the approach and
capture operations and is defined as an area, $A$, enclosed by
a circle centered on the capture interface.
The threshold value is defined using the combination of \gls{esa}['s] recommendations
on the mechanical clearance of mechanisms \citep{ECSS2019a} and
the value of a maximum achievable precision of a typical \gls{gnc} system, in all three axis, during a berthing maneuver (\ie
\SI{0.1}{\meter}), using the following formula $ A = \pi (3 \times 0.1)^2 $ \citep{Jankovic2016b,Fehse2008-Ch7}.
The smaller the interface clearance, the more precise the capture maneuver
needs to be. Therefore, the smaller the interface clearance of a target, the higher its degree of uncooperativeness is.

Using the defined traits, the most cooperative targets are identified
as those having: stable to low tumbling attitude regimes and a ``dedicated''
grapple feature. The most uncooperative targets
are defined instead as those having fast tumbling attitude regimes,
irrespective of other features.

The ADR capture method(s) selection for a particular target, \ie the classification of the associated ADR capture method(s) instance(s), is performed by applying the
class axioms detailed in \cref{tab:capture-methods-restrictions} to the \textsf{ADR\_Capture\_Method} subclasses.

As in case of the definition of characteristics of target objects, these axioms are outlined based on the perceived capabilities of the current state-of-the-art capture methods. Thus, they are provided as a template within the developed ontology. Users are encouraged to modify them or even introduce new ones using software tools presented in the next section.

\begin{sidewaystable}
	\centering{}\caption{\label{tab:capture-methods-restrictions}Class axioms
	used for the classification of ADR capture methods (N/A = not applicable; RB = rocket body; PL = payload).}
	\begin{tabular*}{1\textheight}{@{\extracolsep{\fill}}>{\raggedright}m{0.15\textheight}l>{\raggedright}m{0.05\textheight}>{\raggedright}m{0.05\textheight}>{\raggedright}m{0.15\textheight}>{\raggedright}m{0.05\textheight}ll}
		\toprule
		\multirow{3}{0.15\textheight}{Capture method \\
			subclasses} & \multicolumn{7}{c}{Object properties}\tabularnewline
		\cmidrule{2-8} \cmidrule{3-8} \cmidrule{4-8} \cmidrule{5-8} \cmidrule{6-8} \cmidrule{7-8} \cmidrule{8-8}
		& \multirow{2}{*}{Object type} & \multirow{2}{0.05\textheight}{Breakup \\
			criticality} & \multirow{2}{0.05\textheight}{Passiv. \\
			state} & \multirow{2}{0.15\textheight}{Attitude \\
			regime} & \multirow{2}{0.05\textheight}{Grapple feature} & \multicolumn{2}{c}{Capture interface}\tabularnewline
		\cmidrule{7-8} \cmidrule{8-8}
		&  &  &  &  &  & Material & Clearance\tabularnewline
		\midrule
		\midrule
		\textsf{Manipulator\_Based} & N/A & Low & N/A & Stable-Medium tumbling & True & N/A & N/A\tabularnewline
		\midrule
		\textsf{Clamp\_Based} & RB & Low & N/A & Stable-Medium tumbling & False & N/A & Broad\tabularnewline
		\midrule
		\textsf{Net\_Based} & N/A & Low-Medium & N/A & Stable-Medium tumbling & N/A & N/A & N/A\tabularnewline
		\midrule
		\textsf{Harpoon\_Based} & PL & N/A & True & Stable-Slow tumbling & False & Isotropic & N/A\tabularnewline
		\midrule
		\textsf{Plume\_Impingement} & PL & Low-Medium & N/A & Fast tumbling & N/A & N/A & N/A\tabularnewline
		\midrule
		\textsf{Electromagnetic\_Based} & RB & Low-Medium & N/A & Fast tumbling & N/A & N/A & N/A\tabularnewline
		\midrule
		\textsf{Ablation\_Based} & PL & N/A & True & Fast tumbling & N/A & N/A & N/A\tabularnewline
		\midrule
		\textsf{No\_Solution} & N/A & High & False & Fast tumbling & N/A & N/A & N/A\tabularnewline
		\bottomrule
	\end{tabular*}
\end{sidewaystable}

From the outlined axioms it can be deduced that \textbf{manipulator-based
methods} are designated as a preferred solution for capturing objects (irrespective of their type) having: a low breakup criticality,
stable to medium tumbling attitude and ``dedicated'' grappling feature.
A non-existing grappling feature would make any usage of a manipulator-based
capture method more complex and less safe. Therefore, in these cases
the most suited capture methods are either \textbf{clamp- or tether-based}, selected on the basis of object type, capture interface material and required clearance.

\textbf{Clamp-based solutions} are deemed suitable only for rocket bodies, due to their expected lack of appendages that would otherwise complicate capture maneuvers.

\textbf{Harpoon-based solutions}, in contrast, are associated with
payloads, due to the expected higher efficacy of these methods on
flat surfaces rather than on the curved ones, commonly found on rocket bodies.

Finally, contactless solutions are coupled exclusively with
objects having high levels of uncooperativeness, since any capture
effort using the previously mentioned methods would be considered
very difficult and expensive (in terms of fuel). The selection between
different contactless methods is made on the basis of the estimated
breakup criticality of objects, \ie on the ``required'' stand-off
distance that each method requires to achieve a successful ``capture'' maneuver.
Furthermore, \textbf{plume impingement} and \textbf{ablation-based methods} are
deemed more suitable for payloads, considering that the efficacy of both is
maximized on flat surfaces rather than on the curved ones \citep{Ortiz2017}.
However, payloads usually present a small percentage of conductive material,
with respect to their overall mass, making them challenging targets for
electromagnetic-based methods \citep{Ortiz2017}. As a consequence,
\textbf{electromagnetic-based methods} are bound, in this work, to rocket bodies, usually containing a large percentage of conducting material with respect to their overall mass.

\textbf{No solution} has been found suitable for targets having: a high breakup
criticality, fast tumbling attitude regime and non-passivated state.
The reason behind this result arises from the current unavailability of ADR
capture methods that could safely tackle targets having those characteristics.
%!TEX root = main.tex

\section{Ontology Implementation} \label{sec:ontology_implementation}

The domain-ontology developed within this research is implemented
within the \gls{tracer} repository
hosted on \href{https://zenodo.org/}{Zenodo}\footnote{\url{https://zenodo.org/}}
and \href{https://github.com/}{GitHub}\footnote{\url{https://github.com/}}
platforms. At the moment, the accessibility to the repository on both hosting
platforms is restrained.
However, our near future goal is to provide open access to it, under the
new BSD license, in order to simplify and encourage further development
of the library.

The development workflow of \gls{tracer} is divided into two processes: the
ontology database generation and ontology implementation.
The former includes activities allowing collection
and pre-processing of raw structured and unstructured data, using
the Python programming language.
The latter consists instead of activities leveraging
the output of previous tasks to implement the methodology, described
in \cref{sec:methodology}, employing the ontology editor Protégé.
Moreover, the process includes tasks to import
the desired data into \gls{tracer} and perform unitary tests to assure ontology
consistency.

\subsection{Data sources and pre-processing}\label{subsec:data-sources}

The data sources considered in \gls{tracer} are both, structured and unstructured,
due to the current nonexistence, to best of our knowledge,
of a single source of data containing all the information required for \gls{adr}
capture methods selection, defined in \cref{sec:methodology}.
\gls{esa}['s] \gls{discos} database comes very close with respect to the required amount of
data and provides a machine-to-machine interface:
the \href{https://discosweb-api.sdo.esoc.esa.int/}{DISCOSweb \gls{api}}\footnote{\url{https://discosweb-api.sdo.esoc.esa.int/}},
through which retrieval of structured object data is possible. For
this reason, \gls{tracer} is developed to ``interface'' with the
output of the DISCOSweb \gls{api}, although a manual query to the maintainers of the database may be required due to the existing access
restrictions and limited default information available through the \gls{api} itself.

The structured data expected by \gls{tracer} consists of: the identification properties of potential \gls{adr} targets, their physical dimensions, orbital properties, launch and reentry dates (if any), activity status, onboard propellant type, launcher name and country of origin.

Additional information needs to be provided manually using unstructured sources
that might consist of: web resources, such as
Encyclopedia Astronautica \citep{astronautix2019}, Gunter's Space Page \citep{gunter2019},
Earth Observation Portal \citep{eoportal2019}, RussianSpaceWeb.com
\citep{russianspaceweb2019};
user's manuals/guides of launchers, such as Ariane 5
\citep{arianeV2019}, Atlas V \citep{atlasV2019}, etc.; academic publications in the fields of space debris monitoring and modeling
\citep{Linder2015,Schildknecht2016,Schildknecht2017,Silha2017}.

The unstructured data expected by \gls{tracer} consists of: the attitude states of potential targets, their passivation states, onboard fuels, buses or propulsion unit types, grappling feature existence and potential capture interface properties.

Once the required data is extracted from the chosen sources, the pre-processing
step of the workflow involves integrating that data and creating
a database of the ontology individuals compatible
with \gls{tracer}. The integration of data is done via a custom script written
in Python 3 (version 3.6.8) programming language. The script leverages
the Python's \texttt{pandas} and \texttt{numpy} libraries
(versions 0.24.2 and 1.16.4, respectively) to import, merge, format
and manipulate the extracted data accordingly.
For the computation of breakup probabilities, the Python library
\texttt{lifelines} \citep{davidson_pilon2019}
(versions 0.21.0) is used within the
mentioned script. The \texttt{lifelines} module, built on top of the
\texttt{pandas} library, implements in Python the survival analysis
and provides an \gls{api} to effectively estimate the probability of an event
based on historical data.

\subsection{Data import using \texttt{Owlready2} Python library}\label{subsec:owlready2-data-import}

The next step in the implementation consists of importing data into the domain-ontology, while minimizing the required user input and therefore the possibility of human error.
For this purpose, the experience accumulated within the \gls{ric}--DFKI GmbH, in the field of knowledge representation of the robotics domain \citep{Yueksel2017, Eich2010}, is harnessed and a
method for the manipulation of semantic models, currently being developed within the
D-Rock\footnote{\url{https://robotik.dfki-bremen.de/en/research/projects/d-rock.html}} and Q-Rock\footnote{\url{https://robotik.dfki-bremen.de/en/research/projects/q-rock.html}} projects of the institute, is employed.
The method allows transparent manipulation of
ontologies using the Python 3 programming language and an ontology programming
interfaces, \texttt{Owlready2} (version 0.18)~\citep{Lamy2017}.
\texttt{Owlready2} module allows to load \gls{owl} ontologies as Python objects,
manipulate them, perform reasoning via \texttt{HermiT} or
\texttt{Pellet} semantic reasoners, and save them. It uses
similarities between object models and ontologies to enable high level access
to \gls{owl} ontologies via Python notation, therefore allowing an easy-to-use and
highly concise syntax.
However, \texttt{Owlready2} currently supports a limited
functionality when it comes to knowledge inference compared to other tools,
such as the Protégé editor (introduced in the next subsection). For this
reason, it is not considered in \gls{tracer} for classification purposes.

\subsection{Ontology implementation with Protégé desktop}

Protégé is an open-source framework providing users with a suite of tools to develop, edit
and manage domain models. It fully supports the Semantic Web standards
of the \gls{w3c}, such as the \gls{rdf} and  \gls{owl}~\citep{owl2-2012,Musen2015}.
Furthermore, it allows a connection to description logic reasoners (\ie semantic reasoners) that might be used to probe an ontology for inconsistencies
as well as to infer new knowledge.
Finally, the program provides an option to display human-readable inference explanations, useful to
verify justifications of inferred classifications \citep{protege-desktop2019}.
In view of these considerations, the Protégé Desktop (version 5.5.0)
is selected in this research as the tool of choice for the implementation,
testing and visualization of results of \gls{tracer}.

The particularities of \gls{tracer} implementation consist of:
\begin{inparaenum}[a)]
	\item a minimal hierarchy of classes,
	\item an \gls{odp} usage,
	\item a differentiation between space debris objects and their key characteristics, in terms of individuals.
\end{inparaenum}

The minimal classes hierarchy is intentional in order to reduce the user input and leverage instead the capability of a semantic reasoner to classify the ontology autonomously (\ie to compute the class hierarchy from the imposed axioms).

The \gls{odp} used in \gls{tracer} is the Value Partition.
It is considered as a ``good practice'' in ontology development since its usage allows more robust, cleaner and easier to maintain ontologies~\citep{odp-manchester2009}.
The pattern addresses the problem of representing a descriptive feature
of an object with a constrained set of possible
values (also known as feature space)
\citep{Rodriguez-Castro2012,odp-manchester2009}.
An example in our case can be the ``breakup criticality'' of a space
debris object, which can be regarded as a feature to be
represented and its values ``high'', ``medium'' and ``low''
as its feature space. This way attributes are separated from
elements being described, enforcing an easier to maintain modeling
\citep{odp-manchester2009}.

Lastly, the design choice was made to differentiate between space debris
object themselves and their key attributes, in terms of individuals, similarly to what was done in~\citep{Cox2016,Rovetto2016,Liu2016}.
The relationship with a parent object is guaranteed via object properties.
In this manner, six additional individuals/instances are created
for each space debris object describing its:
\begin{inparaenum}[1)]
	\item attitude,
	\item orbit,
	\item breakup criticality,
	\item probability,
	\item severity,
	\item capture method.
\end{inparaenum}
Two further individuals/instances are used to represent the platform type (\ie
bus or propulsion unit, in case of a payload or rocket body,
respectively) and onboard propellant of an object,
allowing their re-use, in case of objects having the same platform
and/or onboard propellant.
Consequently, a total of eight individuals/instances
are used in \gls{tracer} to compliment the data properties expressed within the space
debris object instance.

The advantages of such an implementation, especially when used in
combination with Protégé, comprise of an easy to use and maintain ontology
in which the characteristics of an object are represented as data properties
while the contextual information and relationship with other individuals
as object properties. Furthermore, the deductive reasoning capabilities
of an ontology can be extended, if required, via the \gls{swrl}~\citep{swrl-2004}, for which Protégé provides
a development environment. This way, the \textsf{orbit\_age} of an
object, for example, can be deducted by a reasoner, based on the
defined \gls{swrl} rules, instead of being derived in the pre-processing step. However, the provided \gls{swrl} development environment was found to be very limited in terms of debugging capabilities and was therefore dismissed in favor of a more familiar Python environment, as illustrated in the previous subsections.

The disadvantages of the described implementation are mainly tied
to ``limitations'' of the Protégé editor itself which requires a
quite pedantic and repetitive insertion of class restrictions, individuals
and properties, without the possibility to automatize the process within
the editor itself. For this reason, the \texttt{Owlready2} is used
to populate \gls{tracer} with the required data and might be used in the future
to manage the developed ontology in more depth, using a more familiar Python
environment.
%!TEX root = main.tex

\section{Ontology Application} \label{sec:ontology_application}

\begin{figure}
	\begin{centering}
		\includegraphics[width=1\linewidth]{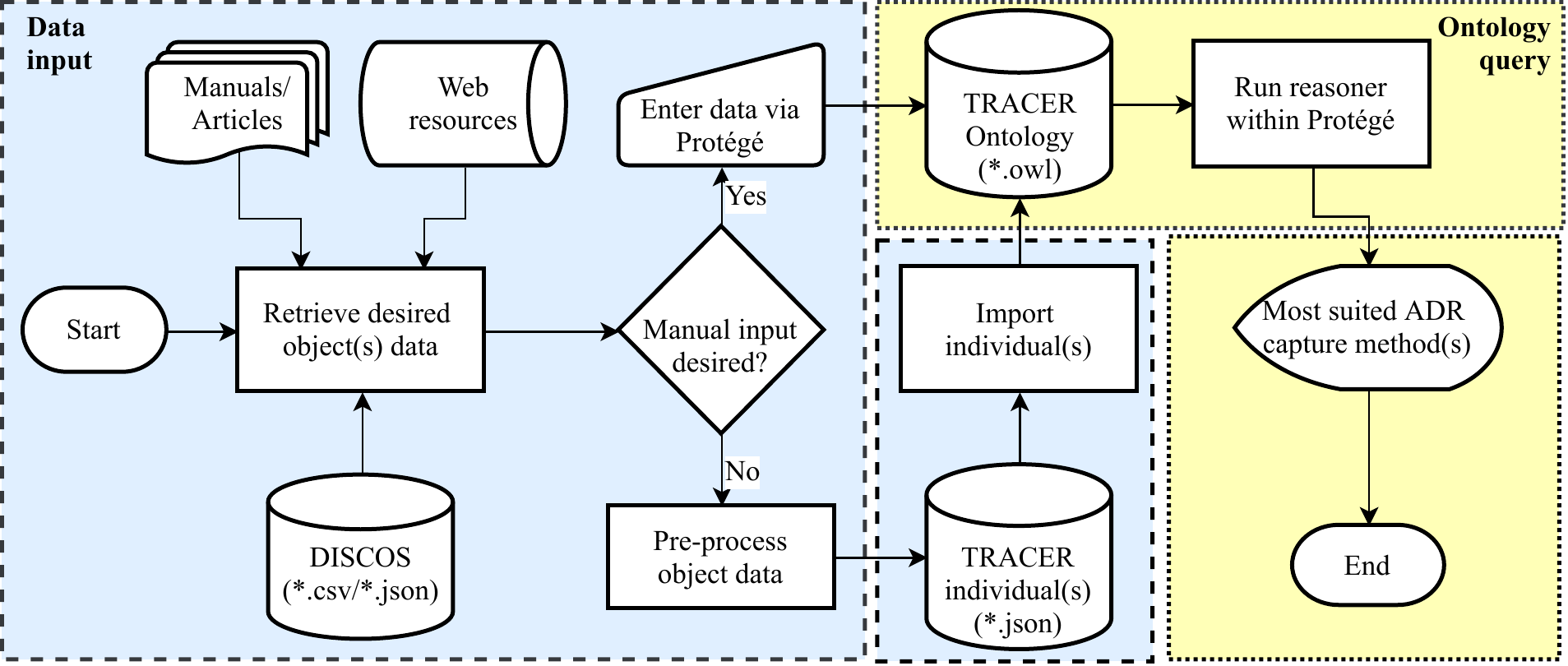}
		\par\end{centering}
	\caption{\label{fig:tracer-usage-workflow}Usage workflow of TRACER (see \protect\href{https://www.iso.org/standard/11955.html}{ISO 5807:1985}
		for symbols definition)}
\end{figure}

The effectiveness of the developed ontology was tested
by applying it onto a database of representative objects. The employed
application workflow is illustrated in \cref{fig:tracer-usage-workflow}
and consists of two main processes: the data input and
ontology query. The former encompasses the collection, pre-processing
and input of desired data into \gls{tracer}, while the latter involves
the knowledge inference from \gls{tracer}, via a software reasoner and query
of results via a GUI.

\subsection{TRACER example database}\label{subsec:tracer-example-database}

The example database created for the evaluation of \gls{tracer}, consists
of 210 ontology individuals, related to 30 large intact cataloged objects (\ie 19 payloads and 11 rocket bodies), for which we were able to obtain information about their attitude states, while avoiding the inclusion of objects having identical properties (\ie bus/propulsion platform and angular rate).

The database was assembled using the available data from \gls{discos}
and the unstructured data sources mentioned in \cref{subsec:data-sources}.
Only one of the considered objects has at the time of writing\footnote{July
2019.} an active status, \ie having the international designator (COSPAR ID) 2014-037A. The rest, either have an unknown
status or can be considered inactive, being deactivated on
purpose or having experienced a catastrophic failure which forced
their early end-of-life disposal.
The characteristics of the first 10 objects of the database are illustrated
in \cref{tab:characteristics-databse-objects}.

\begin{table}
	\begin{centering}
		\caption{\label{tab:characteristics-databse-objects}Characteristics of first
			10 objects of the database (RB = rocket body; PL = payload; LEO = low Earth orbit; MEO = medium Earth orbit; GTO = geostationary Earth transfer orbit).}
		\begin{tabular*}{1\linewidth}{@{\extracolsep{\fill}}>{\centering}m{0.14\linewidth}>{\centering}m{0.1\linewidth}>{\centering}m{0.1\linewidth}>{\centering}m{0.1\linewidth}>{\centering}m{0.1\linewidth}>{\centering}m{0.12\linewidth}>{\centering}m{0.12\linewidth}}
			\toprule
			COSPAR ID & Type & Orbit & CN & Passiv. state & Angular rate\\
			(deg/s) & Orbit age\\
			(years)\tabularnewline
			\midrule
			1978-018B & RB & LEO & 3 & False & 67.8 & 41.96\tabularnewline
			1978-121A & PL & LEO & 6 & False & 2 & 41.11\tabularnewline
			1989-001B & PL & MEO & 6 & False & 38.88 & 31.06\tabularnewline
			1990-005H & RB & LEO & 3 & True & 0 & 30.03\tabularnewline
			1990-045A & PL & MEO & 6 & False & 38.96 & 29.71\tabularnewline
			1991-084C & RB & GTO & 6 & False & 1.74 & 28.13\tabularnewline
			1992-052A & PL & LEO & 6 & True & 32.1 & 27.48\tabularnewline
			1993-061A & PL & LEO & 6 & False & 2 & 26.36\tabularnewline
			1994-021A & PL & MEO & 6 & False & 3.03 & 25.88\tabularnewline
			1994-021B & PL & MEO & 6 & False & 8.41 & 25.88\tabularnewline
			\bottomrule
		\end{tabular*}
		\par\end{centering}
\end{table}

The most represented platforms within the database are the: Uragan Block IIv bus (with five related objects), H10 propulsion unit (with two related objects), Étage à Propergols Stockables (\ie Storable Propellant Stage) (EPS) L9 propulsion unit (with two related objects) and ``ADEOS'' bus (with two related objects).

The distribution of objects within the orbital classes, as defined in \gls{discos}, is the following: \gls{leo}: 15 objects, \gls{meo}: eight objects, \gls{gto}: four objects, \gls{heo}: two objects and \gls{geo}: one object.

The median in-orbit age (since launch) of objects was found to be
\numlist{22.89;19.3} years (with interpercentile ranges being
\numrange{13.7}{25.51} and \numrange{17.55}{25.89} years)
for payloads and rocket bodies, respectively. Therefore, their median breakup
probability values were derived to be equal to
\numlist{3.05e-2;2.55e-2} (with interpercentile ranges
being \numrange{1.35e-2}{3.61e-2} and \numrange{9.86e-3}{3.15e-2}), respectively.

The assumptions made during the consolidation of the database were
the following:
\begin{itemize}
	\item the passivated state of an object was assumed as \textsf{True
	}only where the documentation supporting such a state was found, \eg
	existence of a deactivation date, flight numbers of Ariane launches,
	etc.
	\item the grapple feature existence was assumed as \textsf{True }and
	its material \textsf{isotropic} for all objects within the database,
	considering that they should all be fitted with either a \gls{lar} or a \gls{lva} (or in some cases even
	with an \gls{asap}). Those interfaces
	while might be overall made of a composite material (especially true
	for \glspl{lva}), should all present metallic junction surfaces, considered
	to be suitable as potential grapple features.
	\item the failure date and type information recorded within the database
	refers to a catastrophic type of failure of an object which would
	force its premature shutdown, if at all possible.
\end{itemize}

\subsection{TRACER classification results}\label{subsec:tracer-results}

\begin{figure}
 	\begin{centering}
 		\includegraphics[width=1\linewidth]{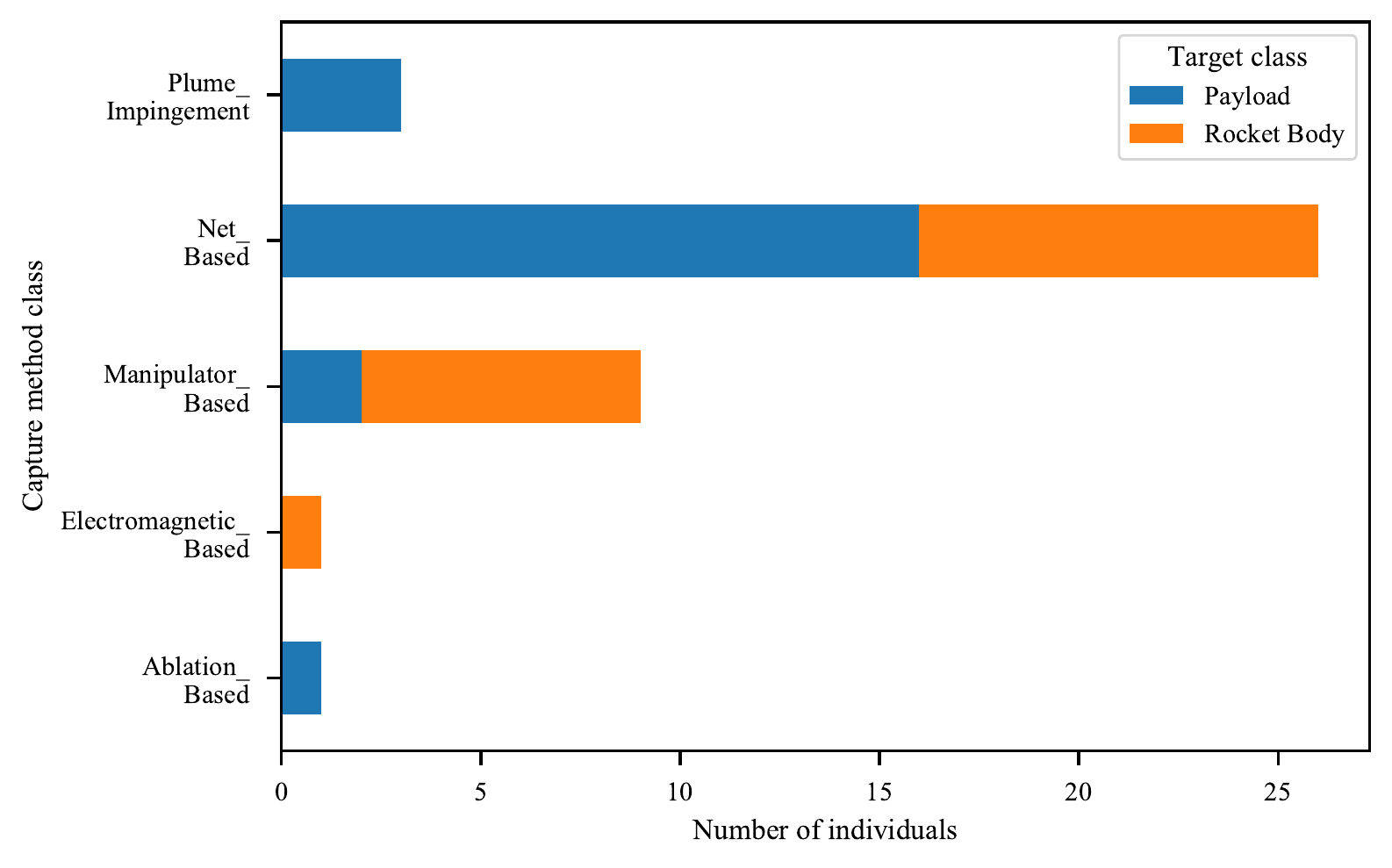}
 		\par\end{centering}
 	\caption{\label{fig:classification-results-per-object-class}ADR capture methods
 		classification results per target object class.}
 \end{figure}

\begin{figure}
	\begin{centering}
		\includegraphics[width=1\linewidth]{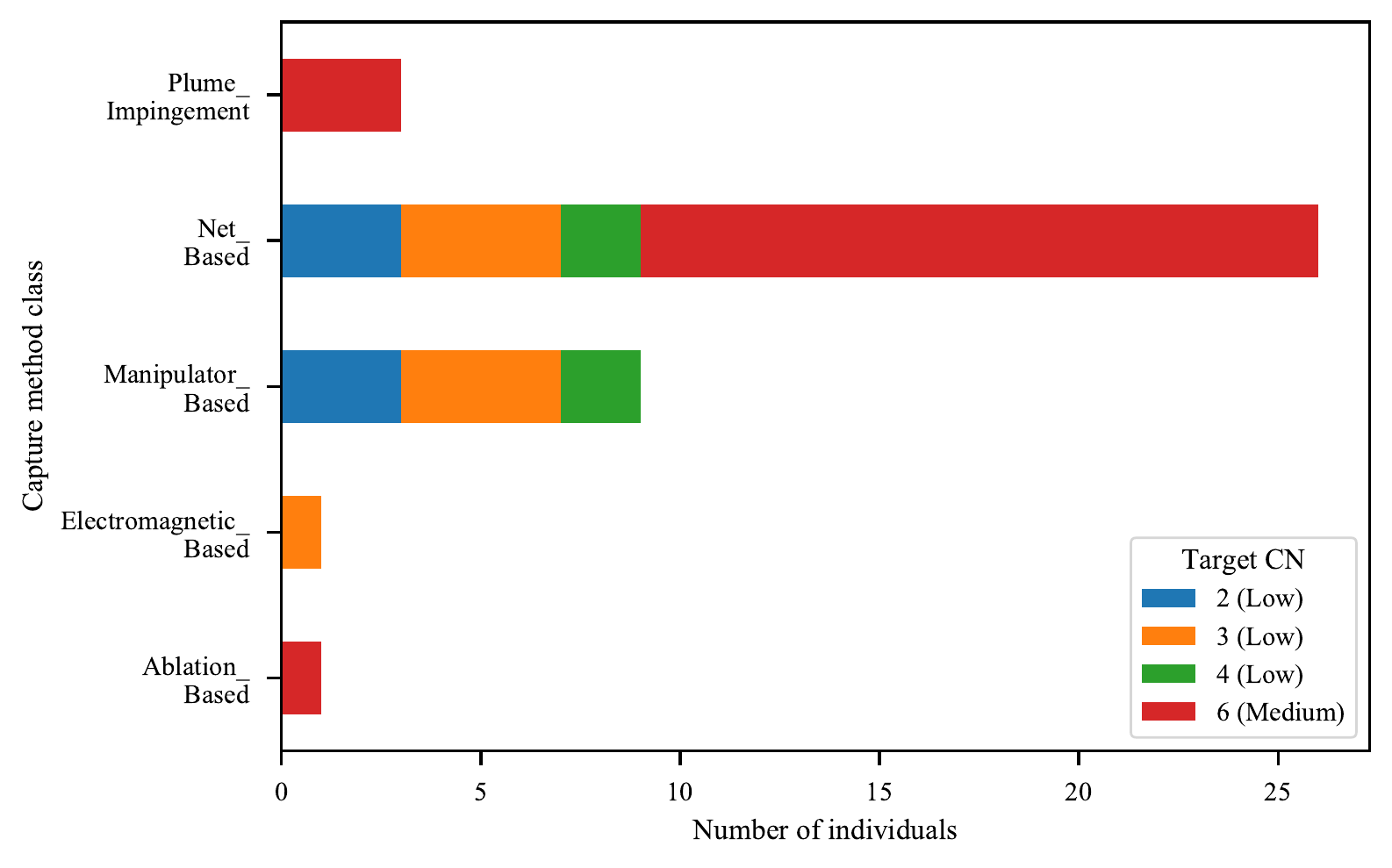}
		\par\end{centering}
	\caption{\label{fig:classification-results-per-criticality-number}ADR capture
		methods classification results per target object criticality number (CN = criticality number).}
\end{figure}

\begin{figure}
	\begin{centering}
		\includegraphics[width=1\linewidth]{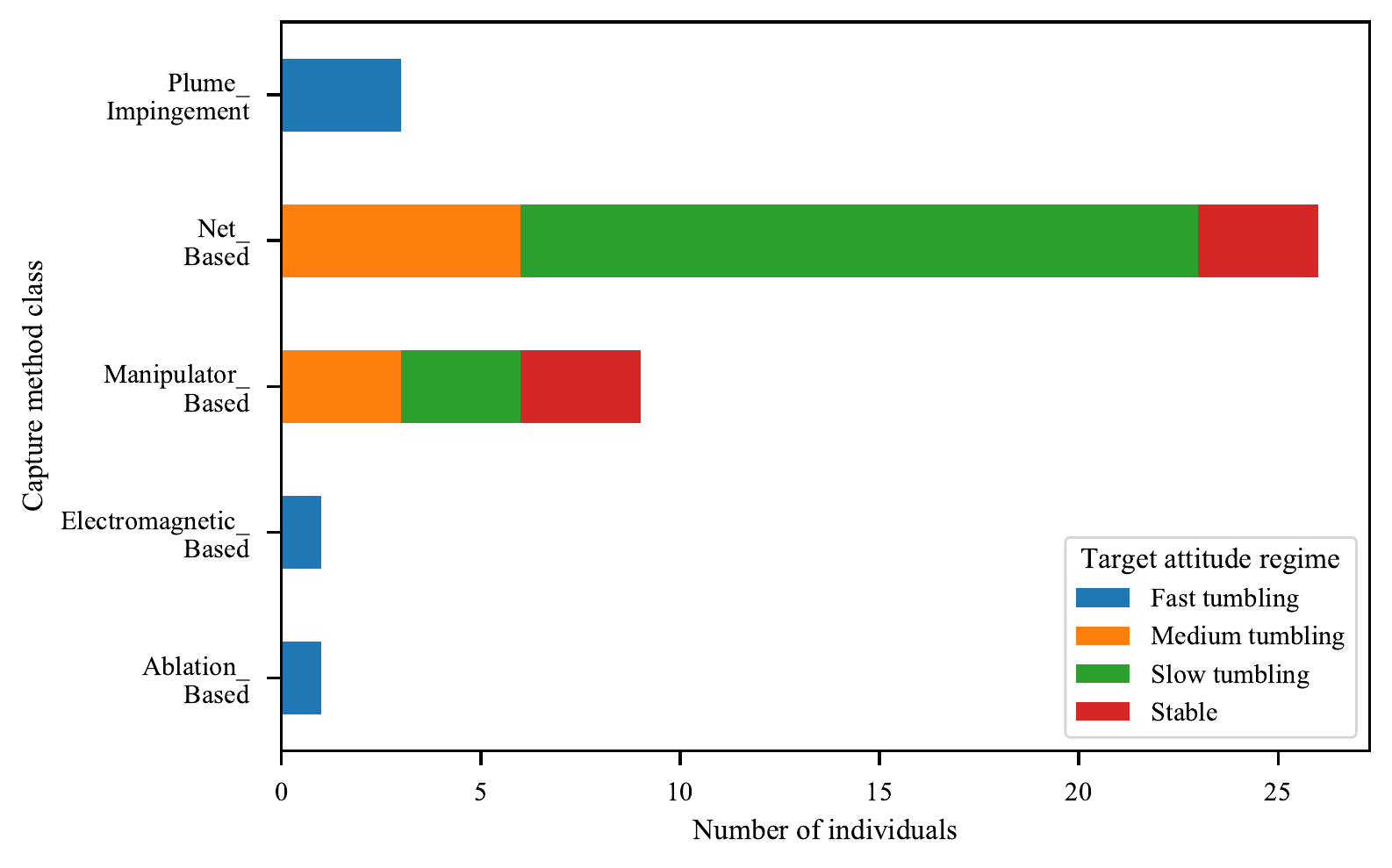}
		\par\end{centering}
	\caption{\label{fig:classification-results-per-attitude-regime}ADR capture
		methods classification results per target object attitude regime (as defined in \cref{subsec:main-classes}).}
\end{figure}

The results of the inferred classification of \gls{adr} capture methods,
related to the \num{210} individuals of the example database, are illustrated in
\cref{tab:classification-results-values} as well as \cref{fig:classification-results-per-criticality-number,fig:classification-results-per-object-class,fig:classification-results-per-attitude-regime}.

\begin{table}
	\caption{\label{tab:classification-results-values}Results of the \gls{adr} capture methods classification with all numeric values being median a part from those within the “No. of individuals” column (CN = criticality number; PN = probability number; SN = severity number).}

	\centering{}%
	\begin{tabular}{>{\raggedright}m{0.25\textwidth}>{\centering}m{0.15\textwidth}ccc>{\centering}m{0.1\textwidth}>{\centering}m{0.1\textwidth}}
		\toprule
		ADR capture methods class & No. of individuals & CN & PN & SN & Angular
		rate\\
		(deg/s) & Orbit age\\
		(years)\tabularnewline
		\midrule
		\textsf{Ablation\_Based} & 1 & 6 & 3 & 2 & 32.1 & 27.48\tabularnewline
		\textsf{Electromagnetic\_Based} & 1 & 3 & 3 & 1 & 67.8 & 41.96\tabularnewline
		\textsf{Manipulator\_Based} & 9 & 3 & 2 & 1 & 2 & 18.32\tabularnewline
		\textsf{Net\_Based} & 26 & 6 & 3 & 2 & 2 & 19.15\tabularnewline
		\textsf{Plume\_Impingement} & 3 & 6 & 3 & 2 & 38.88 & 29.71\tabularnewline
		\bottomrule
	\end{tabular}
\end{table}

The classification was performed using the \texttt{Pellet} semantic reasoner
within the Protégé Desktop, on a 64-bit PC platform equipped with an
Intel\textregistered\ Core\texttrademark\ i7-3630QM CPU, clocked at
\SI{2.40}{\giga\hertz}, and \SI{16}{\gibi\byte} of RAM\@. The required average
time for a classification was found to be \SI{13.5}{\second}.
With lower numbers of individuals, \ie \numlist{142;72} individuals
(corresponding to \numlist{20;10} cataloged objects, respectively), the
average inference computation time was found to be
\SIlist{3.51;1.25}{\second}, respectively.

Overall ten capture methods were classified simultaneously in two classes,
suggesting that, under the considered assumptions, more
than one capture method was found to be suitable for those targets.
The two capture method pairs involved in the simultaneous classification were the following: (\textsf{Manipulator\_Based}, \textsf{Net\_Based}) and
(\textsf{Ablation\_Based}, \textsf{Plume\_Impingement}). The number of
occurrence of these results was nine times and one time, respectively.

\cref{fig:classification-results-per-object-class} portrays
the overall number of capture methods being classified within one
or more \textsf{ADR\_Capture\_Method} subclasses of the ontology,
along with their distribution over the two considered classes of target
objects, \ie payloads and rocket bodies. More specifically,
26 capture methods have been classified within the \textsf{Net\_Based}
class, nine within the \textsf{Manipulator\_Based} class, three within
the \textsf{Plume\_Impingement} class and one within the \textsf{Electromagnetic\_Based}
and \textsf{Ablation\_Based} classes, respectively.

The distribution of inferred classification over the breakup criticality
numbers and attitude regimes of related targets are illustrated in
\cref{fig:classification-results-per-criticality-number,fig:classification-results-per-attitude-regime}, respectively.

%!TEX root = main.tex

\section{Analysis and Discussion} \label{sec:analysis_and_discussion}

The following section analyses the meaning of the results presented in the
previous one and discusses strengths and weaknesses of the developed
ontology, in light of the imposed competency questions defined in
\cref{subsec:ontology-domain-scope}.

\subsection{Results analysis}

The overall classification results (illustrated in \cref{tab:classification-results-values})
point toward the \textsf{ Net\_Based} class as the most numerous one, holding $ \approx $\,\SI{87}{\percent} of all individuals of capture methods. The second most numerous class is the \textsf{Manipulator\_Based},
holding a total of \SI{30}{\percent} of individuals, while the remaining
three classes are found to collect, in total, only \SI{16}{\percent}.
These figures were expected and are to be attributed to: the assumptions
made during the creation of the example database (defined in  \cref{subsec:tracer-example-database})
and restrictions imposed to the \textsf{ADR\_Capture\_Method} subclasses,
(defined in \cref{tab:capture-methods-restrictions}). Indeed,
the database is dominated by slow/medium tumbling objects, having
low/medium breakup criticalities, which makes them ideal targets for
contact-based capture methods, such as nets or manipulators. Furthermore,
all targets were assumed to have a grapple feature that, once
again, made them suitable for manipulator-based methods, especially
in case of objects having low to medium tumbling rates and
low breakup criticalities. Additionally, certain targets were associated with both net- and manipulator-based methods, as their capture methods satisfied more than one class membership constraint. This result is illustrated in
\cref{fig:classification-results-per-criticality-number}, \ie in the mirrored
classification distribution of capture methods associated with targets having
low \glspl{cn} (\ie $ 2 \leq \text{CN} \leq 4 $).

The classification distribution over target object classes, as depicted
in \cref{fig:classification-results-per-object-class}, was
also an anticipated result that can be justified by the overall higher
number of payloads within the example database. In fact, the number
of payloads within the database is almost double with respect to
that of rocket bodies and this is reflected in the overall higher
number of payloads being associated with the most generic capture
method defined, \ie the net-based method. The manipulator-based
method is instead associated with a higher number of target rocket
bodies, due to the higher number of rocket bodies (\ie seven)
with an overall lower median \gls{cn} with respect to that
of payloads.

The association of the remaining classes (\ie \textsf{Plume\_Impingement},
\textsf{Electromagnetic\_Based} and \textsf{Ablation\_Based}) with
one target class or the other was also an awaited outcome that is
a direct consequence of class restrictions specified in~\cref{tab:capture-methods-restrictions}.

The influence of the class restrictions specified in~\cref{tab:capture-methods-restrictions} is particularly evident in~\cref{fig:classification-results-per-criticality-number}
and \ref{fig:classification-results-per-attitude-regime}. Indeed,
the overall characteristics of target objects associated with net-based capture
methods were found to be: low to medium
criticality and stable to medium attitude regime, in line with the
imposed restrictions. In case of objects correlated with manipulator-based methods, those characteristics were identified with: low criticality, stable to medium tumbling attitude regime and the existence of a grapple feature, once
again matching the imposed restrictions of the axioms defined in \cref{tab:capture-methods-restrictions}.
Finally, objects associated with contactless methods
were characterized by low to medium breakup criticalities
and high angular rates, once again as expected.

The total absence of individuals within the \textsf{Clamp\_Based}
or \textsf{Harpoon\_Based} classes reflects the very nature of the
example database, which does not include rocket bodies or payloads with missing
grapple features.

\subsection{Discussion}

The overall scope of this work was twofold, \ie to:
\begin{inparaenum}[a)]
	\item create a framework for data collection, storage and sharing of characteristics of intact derelict objects, using a modern knowledge representation method,
	\item enable a discovery of new domain knowledge, such as the inference of the most suited ADR capture methods of objects.
\end{inparaenum}
The presented results confirm the capability of the developed framework
to fulfill the mentioned objectives, within the constraints defined
in \cref{sec:methodology}. In fact, using the developed framework
a user can capture, in a standardized, formal and machine-interpretable
way, the domain knowledge of \glspl{ido} (in particular of payloads and rocket bodies) useful to \gls{adr}\@. Furthermore, the framework provides a way to represent the degree of hazard of an \gls{ido} to an \gls{adr} capture phase, due to its fragmentation potential and inability to support a capture maneuver.
Additionally, the framework displays how a knowledge representation method can be used to infer the most suited \gls{adr} capture methods, based on the existing
data. Finally, the framework provides tools necessary to transparently
handle the input data from an existing database of space debris into an ontology, a detail often overlooked, or at least not sufficiently documented, by the existing state-of-the-art methods.
However, the required inference time points towards a framework with
an exponential complexity, as demonstrated by the exponentially increasing
computation time with the number of individuals. This suggests the
limitation of the current implementation to deal with a classification
of large number of space objects in one run. Indeed, in case of \num{100}
space objects (which would roughly translate into \num{700} ontology individuals),
the expected computation time\footnote{On the PC platform mentioned in \cref{subsec:tracer-results}.} would require around \SI{90}{\hour}.
Nevertheless, the computation time can be drastically reduced by simply halving
the number of objects to be classified in one run. In fact, considering
\num{50} objects, or circa \num{360} individuals, to be classified within
\gls{tracer} in one run, the expected computation time should be around \SI{4}{\minute}. Hence, this
limitation is not currently seen as a serious impediment of the framework,
especially knowing that the developed repository includes automation
tools adequate to classify large number of \glspl{ido} in batches of \num{50} objects
at a time, for example.
Moreover, in case of large groups of objects to be analyzed, consisting mainly of spacecraft of the same platform/bus (such as in case of planned mega-constellations), objects might be further grouped by attitude regimes (as defined in \cref{subsec:capture-method-selection}) and classification performed only on representative objects of each group, while extrapolating results for the rest. In this manner, the entire group can be analyzed without the need to process each single individual.

Additional limitations of the current version of \gls{tracer} consists of:
\begin{inparaenum}[a)]
	\item its inability to classify all possible types of orbital regimes of parent objects, \eg Extended Geostationary Orbit,
	\item the omission of shape, size and mass properties from characteristics used to determine the degree of uncooperativeness of a target (see \cref{tab:uncoop-characteristics}),
	\item the nonexistence of a cost parameter within the class axioms used for \gls{adr} capture methods classification detailed in \cref{tab:capture-methods-restrictions}.
\end{inparaenum}

Further constraint of the developed framework is its dependence from
the unstructured data that are not always easily retrievable, \eg
attitude states of cataloged objects, their onboard propellant types,
passivation states, etc. Therefore, further research in these areas
is considered of paramount importance, towards a goal of either building
a comprehensive database of objects and their properties, or developing a
machine learning algorithm that could infer the required
data from the existing ones.

Finally, \gls{tracer} in its current form does provide an answer only to
one phase of an \gls{adr} mission, \ie the capture phase.
Other phases, such as the close-range rendezvous or disposal (\eg de-orbit) phases, have not been considered.
Therefore, for \gls{tracer} to represent a comprehensive \gls{adr}
planning solution, something that is out of the scope of the current
article, all possible mission phases of a generic \gls{adr} mission should be
included, \eg as separate ontologies.
%!TEX root = main.tex

\section{Conclusions} \label{sec:conclusions}

The ever-increasing population of space debris has long been recognized by the
scientific community as a critical issue that needs to be addressed with
urgency.
However, choosing the right way to address this issue is currently a difficult task mainly due to the information paradox characterizing the space debris domain.
Past studies have addressed this problem, in the context of \gls{ssa}, with modern knowledge representation techniques, such as taxonomies, ontologies or knowledge graphs.
Nevertheless, none has explicitly addressed the domain of \gls{adr} and most of them appear to overlook the handling of the input data from
existing databases of cataloged objects.
In this work, we bridged this gap by developing a framework, in the form of a domain-ontology, for data collection, storage and sharing of characteristics of intact derelict spacecraft useful to \gls{adr}.
The framework defines the minimal set of physical and dynamical
parameters of an object deemed sufficient to infer, via a semantic reasoner, its most suited ADR capture method(s), safety wise.
This way, not only the management, but also the discovery of the new knowledge is facilitated.
At the same time, the framework is equipped with tools to transparently handle the input of data from an existing space debris catalog, \ie \gls{discos}, thus reducing user input and consequently possibility of a human error.
The practicality and validity of the developed framework were
demonstrated by applying it onto a database of representative objects, for
which we were able to obtain attitude states from publicly available resources.
The overall classification results pointed toward net-based capture methods as the most frequently associated with target objects, followed by manipulator-based methods, as it was expected, considering the nature of objects within the example database and restrictions imposed onto \gls{adr} capture methods classes.

As future work, we expect to further develop the presented ontology to:
\begin{inparaenum}[a)]
	\item address some of its current limitations, such as its inability to classify objects based their shape, size and mass,
	\item add more features, such as its ability to cope with missing data,
	\item extend its domain beyond \gls{adr} capture methods,
\end{inparaenum}
in order to provide the community with a comprehensive tool for \gls{adr} missions planning.

%%%%%%%%%%%%%%%%%%%%%%%%%%%% ACKNOWLEDGMENTS  %%%%%%%%%%%%%%%%%%%%%%%%%%%%%%%%%%

\section*{Acknowledgments}

The research presented in this article was supported in part by the EU Marie Curie Initial Training Network Stardust, FP7-PEOPLE-2012-ITN {[}grant agreement 317185{]}. Therefore, the authors would like to thank the European Commission and the Research Executive Agency for their support and funding. Moreover, this work was conducted using the Protégé resource, which is supported by grant GM10331601 from the National Institute of General Medical Sciences of the United States National Institutes of Health.

%%%%%%%%%%%%%%%%%%%%%%%%%%%%%% REFERENCES %%%%%%%%%%%%%%%%%%%%%%%%%%%%%%%%%%%%%%

%% DO NOT EDIT HERE unless you really know what you are doing!!!
\clearpage
% Define the style of references
\bibliographystyle{elsarticle-num}

% Define reference file
\bibliography{references}

%%%%%%%%%%%%%%%%%%%%%%%%%%% AUTHORS' BIOGRAPHIES  %%%%%%%%%%%%%%%%%%%%%%%%%%%%%%

\clearpage
\section*{Vitae}

% Insert your bio (maximum 100 words) along with a 1 x 1.25in photo

\begin{biography}
	{Marko Jankovic}{figures/MJankovic}
		is a space robotics researcher and project leader of the \href{https://bit.ly/2n7QCjo}{Stardust Reloaded project} at the \href{https://bit.ly/2LDqUgU}{Robotics Innovation Center (RIC) of the German Research Center for Artificial Intelligence (DFKI) GmbH}. His research mainly focuses on the development of solutions for Active Debris Removal and On-orbit Servicing. He was a visiting researcher at the Astronautics Research Group of the University of Southampton and ESA/ESTEC's Advanced Concepts Team. Furthermore, he was a research assistant at the DLR’s Institute of Robotics and Mechatronics. His academic background consists of an ongoing Ph.D. in Engineering, M.Sc. in Astronautical Engineering and B.Sc. in Aerospace Engineering.
\end{biography}

\begin{biography}
	{Mehmed Yüksel}{figures/MYueksel}
	studied Systems Engineering, earning the bachelor and master degrees at the University of Bremen, Germany. He has also earned technician degrees at the High School for Technical Science of the University of Marmara, in the field of Telecomnunications, and at the Technical College Haydarpaşa, in the field of Control Systems in Istanbul, Turkey. He has majors in mechatronics, robotics and system design. He started working at the DFKI GmbH in 2008 in the field of search and rescue robotics. Since 2010, he has been working as a researcher at the DFKI GmbH in the fields of mobility and autonomous driving as well as robotic rehabilitation.
\end{biography}

\begin{biography}
	{Mohammad Mohammadzadeh Babr}{figures/MBabr}
	is an electrical engineer with a specialization in machine learning, and programming. He is a research assistant at the DFKI GmbH since 2018 and his academic background consists of an M.Sc in Automation and Information Engineering, and a B.Sc in Electrical Engineering.
\end{biography}

\begin{biography}
	{Francesca Letizia}{figures/FLetizia}
	is a space debris engineer at IMS Space Consultancy for the European Space Agency (ESA) Space Debris Office, where her main tasks include the assessment of general compliance to space debris mitigation guidelines and of the contribution of a mission to the space debris environment. Previously, she was a mission analysis engineer at OHB Systems, where she worked on trajectory design and collision avoidance strategies for satellites with electric propulsion. Francesca holds a Ph.D. from the University of Southampton, where she graduated with a thesis on space debris clouds evolution and worked on the assessment of the compliance with planetary protection guidelines.
\end{biography}

\begin{biography}
	{Vitali Braun}{figures/VBraun}
	 is a space debris engineer at IMS Space Consultancy for the European Space Agency (ESA) Space Debris Office, where his main tasks include the maintenance of ESA's MASTER model and the DRAMA software suits, which are used in risk assessments and the debris mitigation compliance verification. Vitali holds a Ph.D. from the University of Technology in Braunschweig, where he worked on topics in the space surveillance area and developed methods to provide orbit solutions with predetermined bounded accuracy.
\end{biography}

% END DOCUMENT
\end{document}